\documentclass[sigconf]{acmart}
\AtBeginDocument{%
  }

\copyrightyear{2025}
\acmYear{2025}
\setcopyright{acmlicensed}\acmConference[ICMR '25]{Proceedings of the 2025 International Conference on Multimedia Retrieval}{June 30-July 3, 2025}{Chicago, IL, USA}
\acmBooktitle{Proceedings of the 2025 International Conference on Multimedia Retrieval (ICMR '25), June 30-July 3, 2025, Chicago, IL, USA}
\acmDOI{10.1145/3731715.3733269}
\acmISBN{979-8-4007-1877-9/2025/06}

\usepackage{multirow}
\PassOptionsToPackage{colorlinks=true, urlcolor=pink}{hyperref}
\usepackage{xcolor}

\usepackage{hyperref}
\begin{document}

\title{Advancing Food Nutrition Estimation via Visual-Ingredient Feature Fusion}

%
\author{Huiyan Qi}
\affiliation{%
  \institution{Singapore Management University}
  \country{Singapore}
}
\email{qhyan06040925@gmail.com}

\author{Bin Zhu \dag}
\thanks{$^\dag$ indicates corresponding author.}
\affiliation{%
  \institution{Singapore Management University}
  \country{Singapore}
}
\email{binzhu@smu.edu.sg}

\author{Chong-Wah Ngo}
\affiliation{%
  \institution{Singapore Management University}
  \country{Singapore}
}
\email{cwngo@smu.edu.sg}

\author{Jingjing Chen}
\affiliation{%
 \institution{Fudan University}
 \city{Shanghai}
 \country{China}}
\email{chenjingjing@fudan.edu.cn}

\author{Ee-Peng Lim}
\affiliation{%
  \institution{Singapore Management University}
  \country{Singapore}
}
\email{eplim@smu.edu.sg}

%

\begin{abstract}
Nutrition estimation is an important component of promoting healthy eating and mitigating diet-related health risks. Despite advances in tasks such as food classification and ingredient recognition, progress in nutrition estimation is limited due to the lack of datasets with nutritional annotations. To address this issue, we introduce FastFood, a dataset with 84,446 images across 908 fast food categories, featuring ingredient and nutritional annotations. In addition, we propose a new model-agnostic Visual-Ingredient Feature Fusion (VIF$^2$) method to enhance nutrition estimation by integrating visual and ingredient features. Ingredient robustness is improved through synonym replacement and resampling strategies during training. The ingredient-aware visual feature fusion module combines ingredient features and visual representation to achieve accurate nutritional prediction. During testing, ingredient predictions are refined using large multimodal models by data augmentation and majority voting. 
Our experiments on both FastFood and Nutrition5k datasets validate the effectiveness of our proposed method built in different backbones (e.g., Resnet, InceptionV3 and ViT), which demonstrates the importance of ingredient information in nutrition estimation. \url{https://huiyanqi.github.io/fastfood-nutrition-estimation/}.

\end{abstract}





\maketitle

\section{Introduction}
\label{sec:intro}

Food plays a significant role in promoting healthy living, and related research topics attract increasing attention~\cite{min2019survey}, such as food classification~\cite{bossard2014food, zhu2021learning, chen2024res, gao2024high}, ingredient recognition~\cite{chen2016deep, chen2020zero,liu2024convolution, gui2025efficient}, cross-modal recipe retrieval~\cite{salvador2017learning, zhu2022cross, song2024enhancing}, recipe generation~\cite{salvador2017learning, chhikara2024fire, taneja2024monte, liu2025retrieval}, and nutrition estimation~\cite{thames2021nutrition5k, lo2022intelligent, jiang2024food}.
A key objective of food research is nutrition estimation, which empowers individuals to make healthier dietary choices and mitigate the risk of diet-related diseases~\cite{boswell2018training,tilman2014global}. Various datasets have been constructed in food domain, such as Food-101~\cite{bossard2014food}, VIREO Food-172 and 251~\cite{chen2016deep, chen2020study}, ISIA Food-500~\cite{min2020isia}, and Food2K~\cite{min2023large}, these datasets provide food categories and ingredient labels. In contrast, datasets with nutrition labels are rare due to the much more challenging collection process. Nutrition5k~\cite{thames2021nutrition5k} is a dataset comprising food from cafeterias and includes nutritional annotations, such as calorie and protein, but the data is still limited~\cite{han2023dpf,vinod2022image}. Furthermore, current nutrition estimation methods heavily rely on precise portion size estimation, which is challenging to predict accurately. Additionally, methods based on depth images~\cite{lu2020artificial}, while promising, often rely on datasets with limited variation, making them inadequate for handling complex and diverse dishes, thus limiting their generalization ability.


To address the limitations, we introduce a new dataset, FastFood, which focuses on fast food by collecting food images of popular fast food brands and their corresponding nutritional information. FastFood contains 908 different categories with 84,446 images in total, such as burgers, snacks, beverages, and desserts. In addition, we employ a semi-automatic approach to produce the ingredient labels for each image.
We request GPT-4o to generate ingredients for the provided fast food first, following a manual refinement and correction to ensure the annotation quality. 



In this paper, we further propose a novel model-agnostic Visual-Ingredient Feature Fusion (VIF$^2$) method by integrating image and ingredient features for nutrition estimation. To aggregate ingredient features, we use the text encoder of the pre-trained CLIP model to extract embeddings for all ingredients and align them with the visual feature space through an ingredient projector. Additionally, we enhance robustness during training by leveraging ingredient synonyms replacement and sampling strategies. Ingredient-aware Visual Feature Fusion module is introduced to integrate ingredient features with visual representations extracted from convolutional neural networks (ResNets, InceptionV3) and Transformer-based (ViT) models. Moreover, we utilize separate task-specific heads to regress the nutrition values required for each prediction task. During the testing phase, we use large multimodal models (LMMs) to predict ingredients for test images, refining these predictions through data augmentation and a majority voting mechanism to reduce errors caused by hallucinations. Our proposed method demonstrates that integrating ingredient and visual features can significantly improve nutrition prediction accuracy in our FastFood and Nutrition5k datasets. 

\section{Related Work}
\label{sec:RW}
Food-related datasets play a crucial role in advancing research in this field. Food-101~\cite{bossard2014food} created the first Western food dataset, with 101,000 images categorized into 101 classes. VIREO Food-172 and 251~\cite{chen2016deep, chen2020study} consist of Chinese food and provide ingredient information in addition to category labels. Recipe1M~\cite{salvador2017learning} is a more well-known large-scale food dataset offering images, recipes, and ingredient information. It is commonly used for cross-modal retrieval between recipes and images. Food2K~\cite{min2023large} is another large-scale dataset in the food domain, providing category information for food images to advance scalable food visual feature learning. UEC Food256~\cite{kawano2015automatic} and FoodSeg103~\cite{wu2021large} are two datasets designed for food segmentation tasks. DietDiary dataset~\cite{gui2024navigating} contains sequences of food images captured by users over multiple days, along with corresponding food weights, for the task of weight prediction. Nutrition5K~\cite{thames2021nutrition5k} is widely used for nutrition estimation tasks. It provides food images, videos from different angles, and depth images from two cafeterias, as well as detailed ingredient and nutritional information.
These datasets enable a wide range of research, from food image classification~\cite{liu2016deepfood,mezgec2017nutrinet} to ingredient recognition and recipe retrieval. Nutrition estimation presents greater challenges, requiring accurate recognition of ingredients and their quantities. Early studies, such as Im2Calories~\cite{meyers2015im2calories}, use CNNs to estimate food mass and calories. Recent work highlights the effectiveness of multi-task learning, where models simultaneously predict multiple aspects of nutrition. For instance, ~\cite{ege2017image} introduces a multi-task network for estimating calories, categories, and ingredients. Techniques like integrating RGB-Depth images~\cite{lu2020artificial} and combining object detection with calorie estimation~\cite{ege2019simultaneous} further enhance accuracy.
Meanwhile, Nutrition5K\cite{thames2021nutrition5k} demonstrates the use of depth sensors to improve nutrition prediction. Recently, multimodal large language models~\cite{yin2023foodlmm, jiao2024rode} have also been explored in the food domain to unify various food-related tasks within a single model. 


\begin{figure}
  \centering
\includegraphics[width=0.47\textwidth]{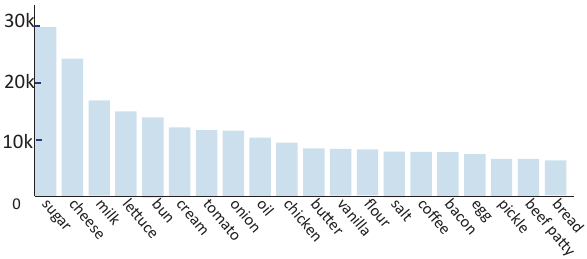} 
\vspace{-0.5cm}
  \caption{Distribution of Top20 ingredients in FastFood dataset.}
  \label{fig:ingredient_frequencies}
\vspace{-0.6cm}
\end{figure}

\begin{figure*}
  \centering
\includegraphics[width=1.0\textwidth]{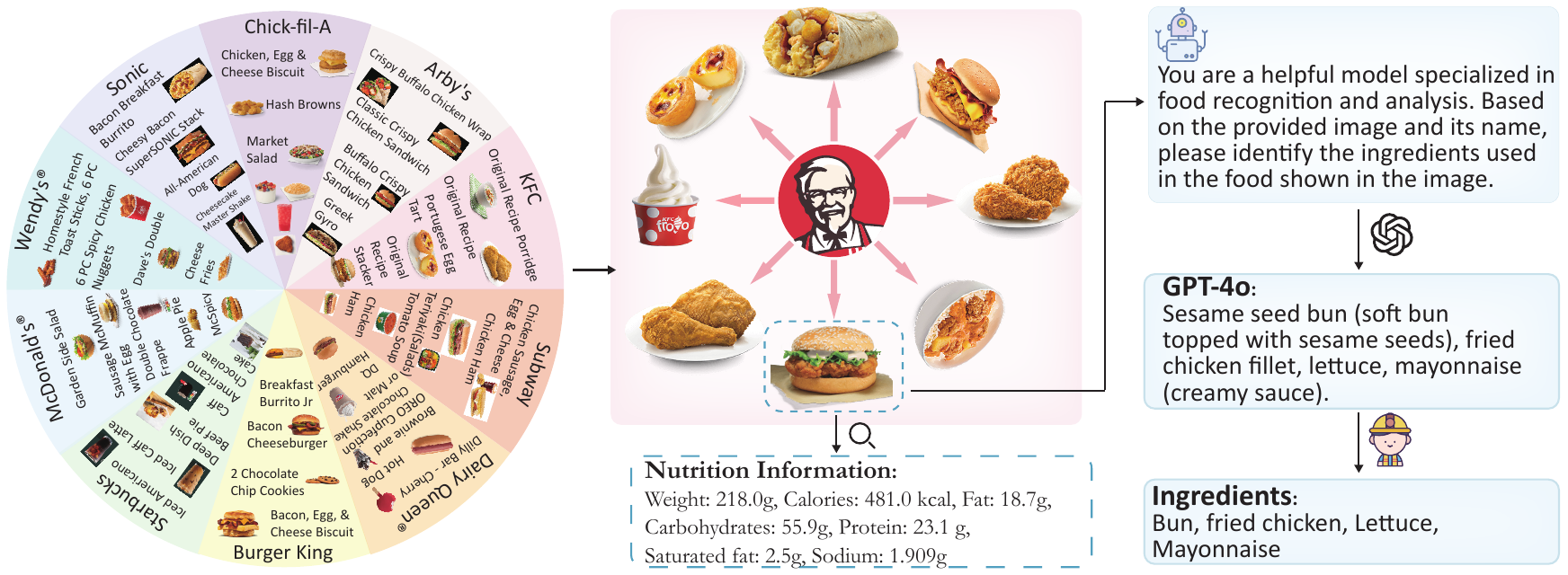} 
  \caption{Overview of our FastFood dataset, highlighting brands, sample images, and annotated ingredients and nutrition.}
  \label{fig:brands}
\end{figure*}

\section{FastFood Dataset Construction}
\label{sec:dataset}
The FastFood dataset is constructed from ten highly renowned fast-food brands worldwide, such as McDonald's and Burger King. We begin by collecting images and nutritional information for each food item directly from the official websites of these fast-food brands. These websites provided high-quality images with clean backgrounds and precise nutritional information, such as calories, fats, carbohydrates, and proteins.
In total, we gathered 908 types of food items, such as bacon cheeseburger and French fries. Figure~\ref{fig:brands} provides an overview of the fast-food brands included and showcases the image examples with ingredients and nutritional information.



\noindent \textbf{Semi-automatic Ingredient Annotation.} We use GPT-4o following a manual correction to derive ingredient annotations in each category of our dataset. The generation process involves providing food images to GPT-4o, accompanied by prompt instructing it to output ingredients relevant to food, while filtering out unrelated terms such as ``straw" or ``fork". While the generated ingredients are generally good, they exhibit issues such as duplicates, vagueness, and context mismatch. Specifically, the generated ingredients often include duplication, such as plural forms. To ensure consistency, these variations are standardized according to a predefined vocabulary.
Additionally, we also remove irrelevant phrases from the ingredients, such as the extra notes within parentheses, such as ``lettuce (200g)", which is revised to ``lettuce".
Vagueness is another issue, where broad terms like ``meat" are used instead of more specific ones like ``beef patty". We manually checked each ingredient to make sure the ingredient annotations were accurate and consistent in our dataset.

Figure~\ref{fig:ingredient_frequencies} showcases the top 20 ingredients in the FastFood dataset, with high-frequency ingredients not only serving as foundational ingredients of fast food but also closely tied to their nutrition characteristics. For example, sugar reflects the high-calorie nature, while cheese and milk are rich in fat and protein. This distribution reflects the impact of high-frequency ingredients on the nutritional composition and health attributes of fast food.

\noindent \textbf{Food Image Expansion.} The images obtained from the official websites are still very limited by numbers and diversity. Thus We expand the dataset via Google image search in two ways:  
(i) Text-Based Image Retrieval: Searching for related images using each food item's name as a keyword. (ii) Image-Based Image Retrieval: Searching for related images using the image of each food item collected from the official website. 
These two methods balance the quality and diversity of the collected images. When searching by food names, the results include images with more complex backgrounds, such as ``a person in a car holding a beef burger".
In contrast, image-based image retrieval typically returns higher-quality images that feature less background and prominently display the food item. After collecting the images via Google search, we manually check each crawled image to ensure the quality of the dataset.

\begin{figure*}[t]
  \centering
\includegraphics[width=1.0\textwidth]{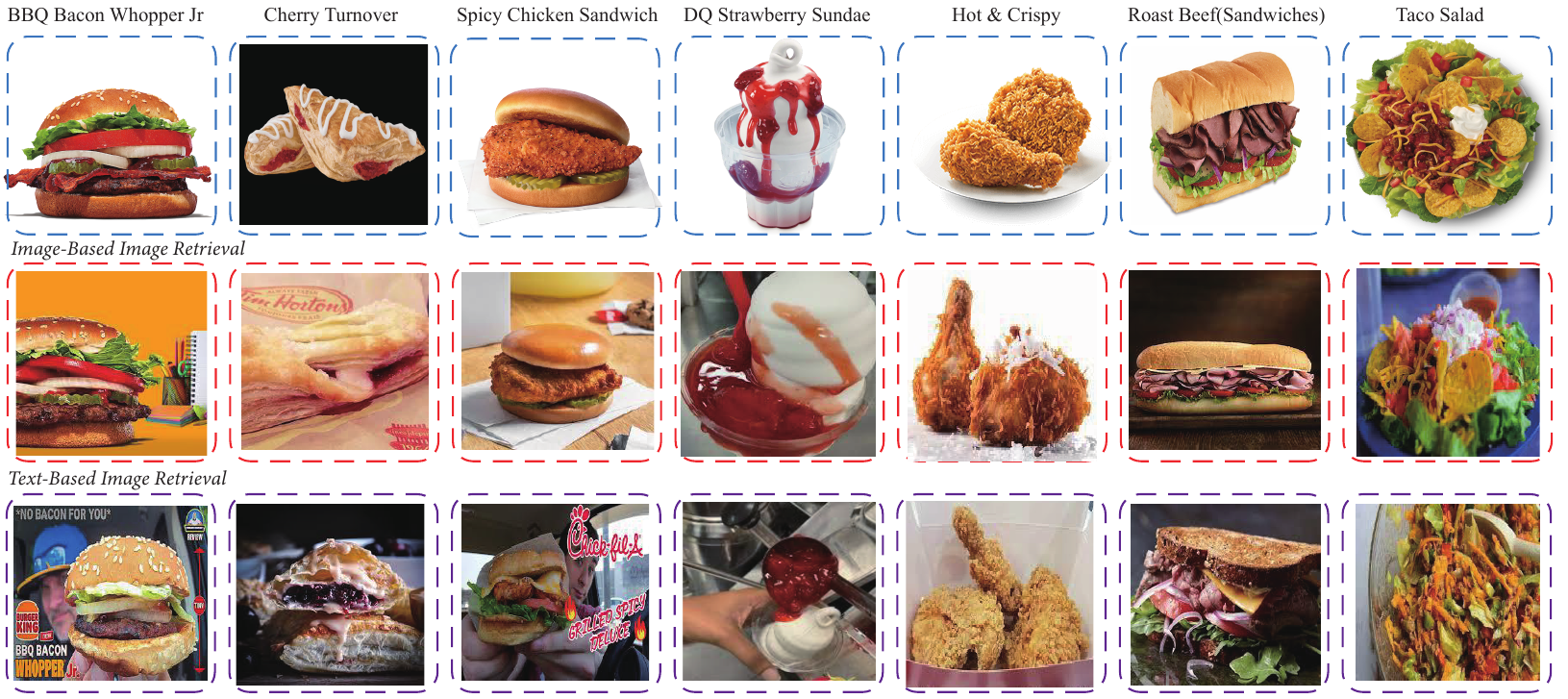} 
  \caption{Data Expansion for Food Items Using Different Retrieval Methods. The first row shows official food item images, while the second and third rows illustrate data expansion achieved through Image-Based Image Retrieval and Text-Based Image Retrieval, respectively.}
  \label{fig:expand}
\end{figure*}

\begin{figure}
  \centering
\includegraphics[width=0.45\textwidth]{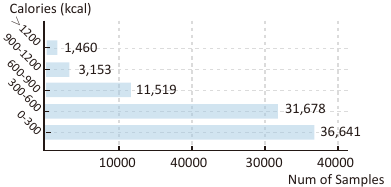} 
  \caption{Distribution of calorie in FastFood dataset.}
  \label{fig:distribution_of_nutrition_content}
\Description{}
\end{figure}

\begin{figure*}[t]
  \centering
  \includegraphics[width=0.98
  \textwidth]{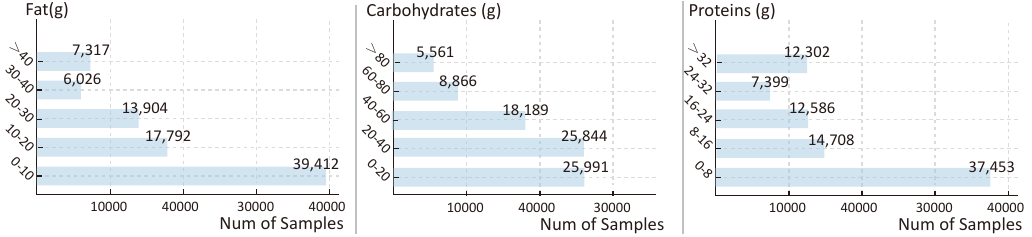} 
  \caption{Distribution of samples in the dataset across different ranges of fat, carbohydrates, and protein content.}
  \label{fig:nutrition_stat}
\Description{}
\end{figure*}

Figure~\ref{fig:expand} shows some examples of images collected using text-based image retrieval and image-based image retrieval. After collection, we exclude images that exhibit blurring, contain multiple food items (except for combo meals), or mismatch the query.
Additionally, we exclude search results where the food size does not match the official image to ensure the accuracy of the nutritional information. For instance, the official website might label a small bottle of milk as fresh milk, but search results using this keyword might include images of large containers of milk.
We review the collected nutritional information for each food item and standardize the units of the same nutritional attributes across different fast food brands.
Figure~\ref{fig:distribution_of_nutrition_content} and Figure~\ref{fig:nutrition_stat} illustrate the distribution of samples in our dataset across different calorie, fat, carbohydrate, and protein ranges, where fat, carbohydrates, and proteins are predominantly in the 0–10 g, 0–20 g, and 0–8 g ranges, respectively, with fewer samples in higher ranges due to the prevalence of single-item food categories in the FastFood dataset, while combo items contribute to the upper ranges.

We eventually obtain 84,446 images, including images obtained from the official websites. We divide these images into training, validation, and test sets in a 70\%, 20\%, and 10\% split, respectively. 




\begin{figure}
  \centering
\includegraphics[width=0.5\textwidth]{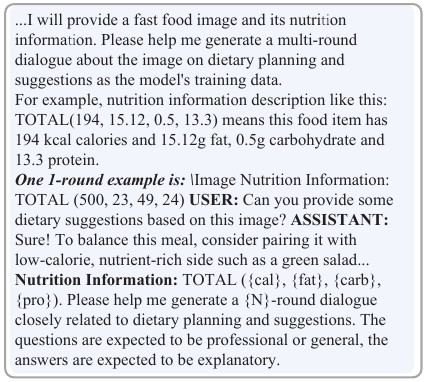} 
  \caption{Prompt template for generating dietary planning and suggestion data.}
  \label{fig:prompt}
\end{figure}

\noindent \textbf{Generating Diet Suggestion dialogue with GPT-4o.}
We selected 5,800 images from our original dataset to generate multi-turn dialogues focused on dietary planning, nutritional analysis, and tailored suggestions using GPT-4o to further show the potential of the dataset. Each image is associated with 2 to 5 rounds of dialogue, resulting in a total of 20,204 dialogue turns, which provide a rich source for training and evaluation of large language models.

The template for generating dietary planning and suggestions data using the GPT-4o API is shown in Figure~\ref{fig:prompt}. We provide GPT-4o with the nutritional content of the food and include an example to ensure the model understands our format. In the template, ${cal}$, ${fat}$, ${carb}$, ${pro}$, and ${N}$ are placeholders that are replaced with the nutritional information and the number of dialogue turns, respectively.
\section{Method}
\label{sec:method}
\begin{figure*}
  \centering
\includegraphics[width=1\textwidth]{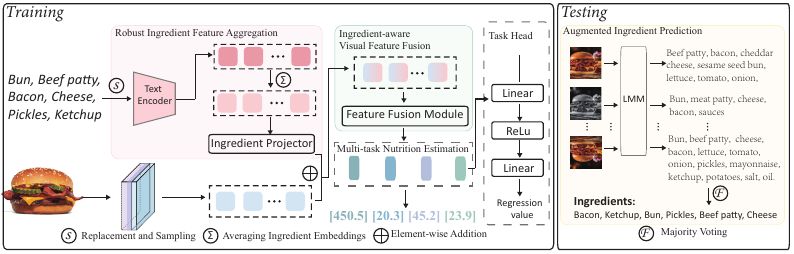} 
  \caption{Overview of the proposed method for nutrition prediction. During training, the ingredient projector aligns robust ingredient features extracted from the ingredient list with visual features. The transformed ingredient features are fused with visual features through the fusion module to produce a comprehensive representation for multi-task nutrition estimation. During testing, a large language model (LMM) is employed to predict ingredients via data augmentation for nutrition estimation.}
  \label{fig:framework}
\end{figure*}

In this section, we introduce Visual-Ingredient Feature Fusion (VIF$^2$), a model-agnostic method that integrates visual and ingredient features to enable accurate prediction of food nutrition. Figure~\ref{fig:framework} illustrates the overall framework of our proposed method. 

\subsection{Robust Ingredient Feature Aggregation}
Given ingredient list $\left\{\mathrm{ing}_{1}, \mathrm{ing}_{2}, \ldots, \mathrm{ing}_{N}\right\}$ for a food image, we employ CLIP's text encoder to each ingredient to generate the corresponding text embedding. 
Then, all the generated embeddings are averaged to obtain the final feature representation of the ingredients $\mathbf{T}_{\text {ingredients }}$ as follows:

\begin{equation}
\label{eq:text_embedding}
\mathbf{T}_{\text {ingredients }}=\frac{1}{N} \sum_{i=1}^{N} \operatorname{CLIP}_{\text {text }}\left(\mathrm{ing}_{i}\right).
\end{equation}

To ensure that the ingredient features and image features are mapped to the same feature space, as shown in Eq.\ref{eq:transform_embedding}, we apply a linear layer and an activation function to $\mathbf{T}_{\text {ingredients }}$, where $\mathbf{W}$ and $\mathbf{b}$ are learnable parameters.
\begin{equation}
\label{eq:transform_embedding}
\mathbf{T}_{\text {transformed }}=\operatorname{ReLU}\left(\mathbf{W} \cdot \mathbf{T}_{\text {ingredients }}+\mathbf{b}\right).
\end{equation}

To make our model more robust for ingredient feature aggregation, we employ GPT-4o to generate a set of synonyms for each ingredient. We ensure that the generated synonyms are semantically similar to the original ingredient name but do not expand into overly general concepts. For example, synonyms for ``lettuce" could be ``romaine lettuce" but not a more generalized term like ``vegetables". This approach aims to preserve the semantic specificity of ingredients while avoiding the introduction of overly vague or unrelated variants. In the training phase, we apply the following two strategies to the ingredient list for each sample: (1) Individual Ingredient Replacement: For each ingredient, we randomly select one of its synonyms with a 50\% probability to replace the original word. This operation enhanced the model's robustness to variations in word forms. (2) Ingredient Sampling: For each image's ingredient list, we sample the entire list with a 50\% probability, randomly selecting a subset as input. 
These two strategies aim to simulate realistic noise in user-provided inputs. In practice, ingredients are often imprecise, incomplete, or expressed in free-form natural language, resulting in inherent ambiguity. By introducing controlled noise during training, we encourage the model to generalize better to such uncertain and variable real-world conditions, where even the "ground-truth" ingredient list may contain ambiguities that are difficult to resolve definitively.

\subsection{Ingredient-aware Visual Feature Fusion}
We propose a model-agnostic ingredient-aware visual feature fusion method to integrate image and ingredient features. We select two well-known convolutional models, ResNet~\cite{he2016deep} and InceptionV3~\cite{szegedy2016rethinking}, as well Transformer-based model ViT~\cite{dosovitskiy2020image} as our baselines.
ResNet and Inception models generate multi-scale feature maps for food images through convolution and pooling operations, transforming the image representation from low to high level~\cite{he2016deep, szegedy2016rethinking}, thus we select intermediate feature maps $\mathbf{X}_{\text {img }}$ to fuse ingredient features. Specifically, for ResNet50 and ResNet101, we choose the convolutional output of block2 for feature fusion, while for InceptionV3, we select the feature map after its second pooling layer.
To incorporate ingredient information, as shown in Eq.\ref{eq:fusion}, the linearly transformed text embeddings $\mathbf{T}_{\text {transformed }}$ are added channel-wise to the feature map $\mathbf{X}_{\text {img }}$, where $\mathbf{T}_{\text {transformed }} \in \mathbb{R}^{C}$ is expanded to match the dimensions of the feature map through a broadcasting mechanism.

\begin{equation}
\label{eq:fusion}
\mathbf{X}_{\text {fused }}=\mathbf{X}_{\text {img }}+\mathbf{T}_{\text {transformed }.}
\end{equation}

In contrast, images are processed through the patch embedding module to generate a series of patch tokens in ViT model. To facilitate better interaction between ingredient information and image features, we introduce linearly transformed ingredient feature as a separate token, which is input into the transformer encoder alongside the class token and patch tokens. Denote the input sequence $\mathbf{X}_{\text {tokens }} = \left[\mathbf{X}_{\mathrm{cls}}, \mathbf{T}_{\text {transformed }}, \mathbf{X}_{\text {patches }}\right]$, where $\mathbf{X}_{\mathrm{cls}}$ represents the classification token used to capture global information, and $\mathbf{X}_{\text{patches}}$ refers to the patch embeddings. We combine the input sequence with positional encoding $\mathbf{P}_{\text {pos }}$ and pass it through the Transformer layers:

\begin{equation}
\label{fusion_embedding}
\mathbf{X}_{\text {encoded }}=\text { Transformer }\left(\mathbf{X}_{\text {tokens }}+\mathbf{P}_{\text {pos }}\right).
\end{equation}
Finally, we use the representation of the class token for subsequent nutrition prediction tasks.

\subsection{Multi-task Nutrition Estimation}
Following Nutrition5k~\cite{thames2021nutrition5k}, we train a separate head for each nutrition (i.e., calories, fats, carbohydrates, and proteins), which is a two-layer fully connected neural network. The first layer projects the input features to a 4096-dimensional hidden layer, and the second layer maps the hidden layer's output to a single scalar value. A ReLU activation function is applied after the first layer to enhance the model's nonlinear representational capacity.

We employ the Mean Absolute Error (MAE) to train the model as in  Eq.~\ref{eq:MAE}, $n_{i}$ represents the number of samples for task $i$, $\hat{y}_{i, j}$ and $y_{i, j}$ are the predicted value and the true value of the $j_{th}$ sample in task $i$, respectively.

\begin{equation}
\label{eq:MAE}
\operatorname{MAE}\left(\hat{y}_{i}, y_{i}\right)=\frac{1}{n_{i}} \sum_{j=1}^{n_{i}}\left|\hat{y}_{i, j}-y_{i, j}\right|
\end{equation}

The total loss of the model is represented as follows:
\begin{equation}
\label{eq:LOSS}
\mathcal{L}=\sum_{i=1}^{N} \operatorname{MAE}\left(\hat{y}_{i}, y_{i}\right).
\end{equation}

\subsection{Augmented Ingredient Prediction}

Acquiring ingredient labels is often challenging in practice. Therefore, during the testing phase, we assume only images are available. Consequently, we utilize off-the-shelf LMMs (e.g., LLaVa~\cite{li2023llava}) to generate the ingredients for these test images. Since hallucination is a common problem with LLMs~\cite{ji2023survey, liu2024survey, tonmoy2024comprehensive, xu2024hallucination}, we notice that the results from current LLMs frequently produce ingredients that do not exist in the image. To address this issue, we introduce augmented ingredient prediction. 
Specifically, for each test image $\mathbf{I}$, a set of data augmentation methods $\left\{\mathcal{T}_{1}, \mathcal{T}_{2}, \ldots, \mathcal{T}_{K}\right\}$ (e.g. Random Rotation, Horizontal Flip, Random Crop with  maintaining at least 70\% of the original size, and Convert to Grayscale) are applied to produce $K$ augmented images: 

\begin{equation}
\label{eq:aug_image}
\mathbf{I}_{k}=\mathcal{T}_{k}(\mathbf{I}), \quad k \in\{1,2, \ldots, K\},
\end{equation}
$\mathcal{T}_{k}$ represents the $k$-th augmentation method.
Subsequently, each augmented image $\mathbf{I}_{k}$ is fed into the LMM to predict the ingredients $\mathcal{S}_{k}$:
\begin{equation}
\label{eq:generate_ing}
\mathcal{S}_{k}=\operatorname{LMM}\left(\mathbf{I}_{k}\right).
\end{equation}

Finally, a majority voting mechanism is employed to filter the ingredients from all the predicted results $\mathcal{S}_{\text{all}}$. We count the frequency of occurrence for each ingredient, and any ingredient with a count exceeding the predefined threshold $\tau$ is included in the final prediction result:

\begin{equation}
\label{eq:vote}
\mathcal{S}_{\text {final }}=\left\{\text { ing } \mid n(\text { ing }) \geq \tau, \text { ing } \in \mathcal{S}_{\text {all }}\right\}.
\end{equation}
The final predicted ingredients and image are then used as input for our model to predict nutrition values.

\section{Experiments}
\label{sec:experiments}
\subsection{Experimental Setup}
\noindent\textbf{Datasets and Evaluation.}
Our experiments are conducted on our FastFood and Nutrition5k~\cite{thames2021nutrition5k} datasets. Nutrition5k provides RGB images, multi-angle videos, and depth images of dishes captured in two cafeterias using sensor devices. It is worth noting that publicly available large-scale datasets with rich nutritional annotations and multimodal data (e.g., RGB, depth, video) are extremely scarce. Nutrition5k is one of the few such datasets, making it a valuable benchmark for evaluating nutrition estimation models. During training, we follow~\cite{thames2021nutrition5k} by sampling every five frames from the video data in the Nutrition5k dataset. After extracting video frames, Nutrition5k contains 230,000 images for training, 5,700 for validation, and 38,000 for testing. In all experiments, we evaluate model performance using MAE and also report the Relative Error Percentage, which expresses MAE as a percentage of the corresponding field's mean value.
Additionally, we evaluate Nutrition5k under two testing protocols. First, we sample one frame for each frame in test videos.
Second, based on the first protocol, we select the optimal frame per video to mitigate the potential redundancy and noise, such as blurriness or distortions.

\par
\noindent\textbf{Implementation details.}
We adopt the widely used architectures, including ResNet50, ResNet101, InceptionV3 and ViT (base16) as backbones for experiments. When InceptionV3 serves as the backbone network, the input image resolution is set to 299x299, which is the standard input size required for InceptionV3. For the other five models, the input image resolution is 224x224. During training, we optimize the models using the RMSProp algorithm with an initial learning rate of 1e-4, momentum of 0.9, decay of 0.9, and epsilon set to 1.0. All models are initialized with the weights pre-trained on ImageNet. Each model is trained for 100 epochs, with a batch size of 64 for the FastFood dataset and 256 for the Nutrition5k dataset. 

\subsection{Performance Comparison}


\begin{table}[htbp]
\caption{Performance comparison between baselines and their counterparts enhanced with our proposed VIF$^2$ on the FastFood dataset.
\label{tab:results_FastFood}}
\vspace{-0.2cm}
\center
\scalebox{0.75}{
\begin{tabular}{lcccc}
\toprule
Method  & Caloric MAE & Fat MAE & Carb MAE & Protein MAE \\
\midrule
Resnet101~\cite{he2016deep} & 118.04 / 29.84\% & 7.06 / 43.98\% & 15.57 / 39.14\% & 7.18 / 40.30\% \\
Resnet50~\cite{he2016deep} & 118.07 / 29.85\% & 6.97 / 43.40\% & 15.32 / 38.52\% & 7.03 / 39.43\% \\
ViT~\cite{dosovitskiy2020image} & 141.39 / 35.76\% & 8.20 / 51.07\% & 18.76 / 47.17\% & 8.58 / 48.17\% \\
InceptionV3~\cite{szegedy2016rethinking} & 142.57 / 36.04\% & 10.51 / 65.49\% & 22.74 / 57.18\% & 11.62 / 65.17\% \\
\midrule
Resnet101 + \textbf{VIF$^2$} & \textbf{61.26 / 15.49\%} & \textbf{3.60 / 22.40\%} & \textbf{8.76 / 22.02\%} & \textbf{3.87 / 21.73\%} \\
Resnet50 + \textbf{VIF$^2$} & 63.86 / 16.15\% & 3.99 / 24.85\% & 9.67 / 24.32\% & 4.18 / 23.44\% \\
ViT + \textbf{VIF$^2$} & 76.62 / 19.38\% & 4.75 / 29.60\% & 12.01 / 30.20\% & 5.29 / 29.71\% \\
InceptionV3 + \textbf{VIF$^2$} & 80.86 / 20.44\% & 4.93 / 30.70\% & 11.02 / 27.72\% & 4.94 / 27.71\% \\
\bottomrule
\end{tabular}
}

\end{table}

\begin{table}[htbp]
\caption{Performance comparison between baselines and their counterparts with our VIF$^2$ on the Nutrition5k dataset. \label{tab:results_nutrition5k}}
\vspace{-0.2cm}
\center
\scalebox{0.73}{
\begin{tabular}{lcccc}
\toprule
Method  & Caloric MAE & Fat MAE & Carb MAE & Protein MAE \\
\midrule
\multicolumn{5}{c}{Testing Protocol 1}\\
Resnet101~\cite{he2016deep} & 103.03 / 42.81\% & 9.94 / 69.35\% & 12.91 / 72.86\% & 7.56 / 56.56\% \\
Resnet50~\cite{he2016deep} & 104.68 / 43.49\% & 9.97 / 69.58\% & 12.92 / 72.91\% & 7.64 / 57.09\% \\
ViT~\cite{dosovitskiy2020image} & 148.47 / 61.69\% & 11.57 / 80.78\% & 12.67 / 71.48\% & 9.49 / 70.98\%\\
InceptionV3~\cite{szegedy2016rethinking} & 106.50 / 44.25\% & 9.93 / 69.33\%  & 13.06 / 73.70\%  & 7.72 / 57.73\% \\

\bottomrule
\multicolumn{5}{c}{Testing Protocol 1} \\
Resnet101 + \textbf{VIF$^2$} & \textbf{83.75 / 34.80\%} & \textbf{6.94 / 48.45\%} & 5.64 / 31.85\% & 3.27 / 24.46\% \\
Resnet50 + \textbf{VIF$^2$} & 84.49 / 35.11\% & 7.01 / 48.95\% & \textbf{5.63 / 31.77\%} & \textbf{3.15 / 23.55\%} \\
ViT + \textbf{VIF$^2$} & 87.31 / 36.28\% & 7.26 / 50.71\% & 5.58 / 31.49\% & 3.24 / 24.19\% \\
InceptionV3 + \textbf{VIF$^2$} & 90.65 / 37.67\% & 7.63 / 53.27\% & 6.57 / 37.05\% & 3.75 / 28.02\% \\
\midrule
\multicolumn{5}{c}{Testing Protocol 2} \\
Resnet101 + \textbf{VIF$^2$} & \textbf{23.32 / 9.69\%} & \textbf{1.93 / 13.50\%} & \textbf{1.57 / 8.87\% }& \textbf{0.91 / 6.82\%} \\
Resnet50 + \textbf{VIF$^2$} & 28.84 / 11.98\% & 2.39 / 16.71\% & 1.92 / 10.84\% & 1.08 / 8.04\% \\
ViT + \textbf{VIF$^2$} & 31.07 / 12.91\% & 2.58 / 18.05\% & 1.99 / 11.21\% & 1.15 / 8.61\% \\
InceptionV3 + \textbf{VIF$^2$} & 31.81 / 13.22\% & 2.68 / 18.69\% & 2.31 / 13.00\% & 1.32 / 9.83\% \\
\bottomrule
\end{tabular}
}
\end{table}

Tables~\ref{tab:results_FastFood} and ~\ref{tab:results_nutrition5k} list the results on the FastFood and Nutrition5k dataset respectively. After incorporating our proposed VIF$^2$, the performances are significantly improved compared with baseline counterparts for both datasets. 
For instance, the best-performing model, ResNet101+VIF$^2$, achieves a Caloric MAE of 61.26—a remarkable 48.1\% reduction from the baseline ResNet101's 118.04. Additionally, the relative error decreases from 29.84\% to 15.49\%. Similarly, on the Nutrition5k dataset, ResNet101+VIF$^2$ achieves the best results, reducing the Caloric MAE to 83.75, an 18.7\% improvement over the baseline ResNet101. It is not surprising that the performance of testing protocol 2 shows much better performance than that of testing protocol 1 as we select the best-performed frame for each testing video. 
In addition, 
our proposed method Resnet101+VIF$^2$ with testing protocol 2 demonstrates a significant improvement over state-of-the-art (SOTA) methods, including Google-Nutrition-depth~\cite{thames2021nutrition5k}, Google-Nutrition-monocular~\cite{thames2021nutrition5k}, RGB-D Nutrition~\cite{shao2023vision}, and Swin-Nutrition~\cite{shao2022rapid}. Among these, DPF-Nutrition~\cite{han2023dpf} achieves the best caloric estimation performance on Nutrition5k using monocular images. Compared to this method, our approach reduces caloric MAE by 38.47\% and protein MAE by 74.44\%, highlighting its effectiveness in food nutrition estimation. Since the evaluation protocol for these methods is unclear, we do not include them in the comparison tables. Under a fair comparison, our proposed method consistently outperforms all baselines.

Table~\ref{tab:results_FastFood-908_gt} provides the performance of different models combined with VIF$^2$ when testing with the ground truth ingredients of FastFood dataset. 
Compared to the baselines presented in the main text, all models achieved significant reductions in nutrition prediction. Among them, ResNet101 + VIF$^2$ performs the best, with a caloric MAE reduced to 44.71 and a relative error of only 11.30\%. Similar to the experiment conducted with ingredients generated by other LMMs, ResNet101 remains the best-performing model, while the other models perform slightly worse than ResNet101.
This result can be regarded as the upper bound of VIF$^2$ under ideal conditions, providing a reference for evaluating the impact of generated ingredients on model performance and demonstrating the general applicability of VIF$^2$ across different architectures.

\begin{table}[htbp]
\caption{Performance comparison of different architectures combined with VIF$^2$ using ground truth ingredients from the FastFood dataset during testing. 
\label{tab:results_FastFood-908_gt}}
\center
\scalebox{0.75}{
\begin{tabular}{lcccc}
\toprule
Method  & Caloric MAE & Fat MAE & Carb MAE & Protein MAE \\
\midrule
Resnet101 + \textbf{VIF$^2$} & 44.71 / 11.30\% & 2.59 / 16.13\% & 7.06 / 17.75\% & 2.85 / 15.99\% \\
Resnet50 + \textbf{VIF$^2$} & 47.39 / 11.98\% & 3.08 / 19.21\% & 8.07 / 20.28\% & 3.21 / 18.02\% \\
ViT + \textbf{VIF$^2$} & 48.29 / 12.21\% & 3.42 / 21.29\% & 9.19 / 23.09\% & 3.41 / 19.12\% \\
InceptionV3 + \textbf{VIF$^2$} & 48.53 / 12.27\% & 3.07 / 19.15\% & 7.77 / 19.53\% & 3.06 / 17.18\% \\
\bottomrule
\end{tabular}
}

\end{table}

\begin{figure}
  \centering
\includegraphics[width=0.47\textwidth]{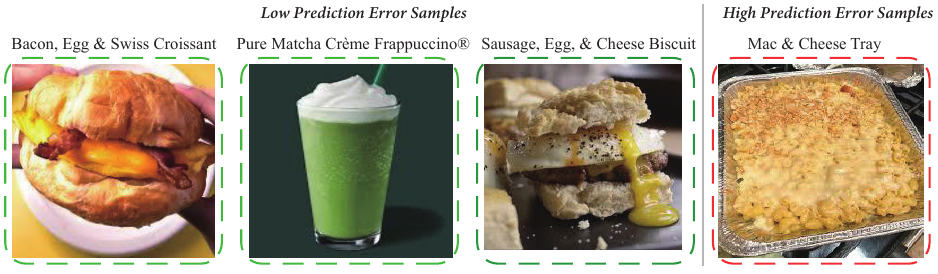} 
  \caption{Examples of typical samples with low and high prediction errors.}
  \label{fig:prediction_samples}
\end{figure}

Figure~\ref{fig:prediction_samples} illustrates typical samples with low and high prediction errors. The three images on the left, with more accurate predictions, demonstrate the model's strong performance in handling foods with clear shapes and distinct ingredients. For instance, the 'Bacon, Egg \& Swiss Croissant' features prominent visual characteristics and well-defined ingredients, making it easy for the model to recognize. In contrast, the image on the right, which shows higher prediction errors ('Mac \& Cheese Tray'), has more complex visual features, highly mixed ingredients, and lacks clear boundaries, making it difficult for the model to extract key features effectively. This highlights the challenges the model faces when dealing with foods that exhibit complex visual characteristics and diverse ingredient compositions.
These results clearly demonstrate the effectiveness of visual-ingredient feature fusion in enhancing nutrition estimation performance. Furthermore, our findings indicate that convolutional networks, such as ResNets and InceptionV3, outperform transformer-based models like ViT in nutrition-related tasks~\cite{thames2021nutrition5k}.

\subsection{Ablation Study}
\begin{figure}
  \centering
\includegraphics[width=0.42\textwidth]{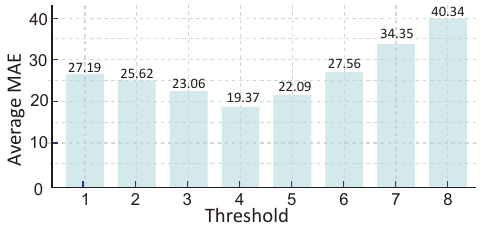} 
  \caption{The average MAE across four tasks as ResNet101 integrates ingredient features generated by LLaVA1.6 under different thresholds.}
  \label{fig:threshold}
\end{figure}

\begin{table}[h]
\vspace{-0.2cm}
\caption{Ablation study by integrating ingredient features into different layers of ResNet101 and InceptionV3.\label{tab:ablation_layer}}
\center
\scalebox{0.55}{
\begin{tabular}{ccc|ccc}
\toprule
model & Injection layer & Average MAE &  model & Injection layer & Average MAE \\
\midrule
& block1 & 22.20/24.76\% & & PostMaxPool2 & 27.10/28.26\% \\
& block2 & \textbf{20.47/22.22\%} & & Mixed6e (w AUX) & \textbf{25.44/26.64\%} \\
& block3 & 21.57/22.20\% & & Mixed6e (w/o AUX) & 25.66/26.77\% \\
\multirow{-4}{*}{\textbf{Resnet101~\cite{he2016deep}}} & block4 & 33.09/32.94\% & \multirow{-4}{*}{\textbf{InceptionV3~\cite{szegedy2016rethinking}}} & PostMixed7c & 32.75/33.55\% \\
\bottomrule
\end{tabular}
}
\end{table}

\begin{table}[htbp]
\vspace{-0.2cm}
\caption{Performance comparison of different large multimodal models for generating ingredients during testing.  \label{tab:ablation_LMM}}
\vspace{-0.5cm}
\center
\scalebox{0.65}{
\begin{tabular}{lccccc}
\toprule
LMM & Caloric MAE & Fat MAE & Carb MAE & Protein MAE \\
\midrule
LLaVA1.6(7B)~\cite{liu2024improved} & \textbf{61.26 / 15.49\%} & \textbf{3.60 / 22.40\%} & \textbf{8.76 / 22.02\%} & \textbf{3.87 / 21.73\%}\\
Qwen2-VL(7B)~\cite{wang2024qwen2} & 69.56 / 17.59\% & 4.30 / 26.82\% & 10.14 / 25.49\% & 4.54 / 25.45\% \\
LLaVA-NeXT(LLama-3 8B)~\cite{zhang2024video} & 73.10 / 18.48\% & 4.50 / 28.03\% & 10.54 / 26.49\% & 4.63 / 26.00\% \\
\bottomrule
\end{tabular}
}
\end{table}

\noindent \textbf{Ingredient feature injection layer.} In Table~\ref{tab:ablation_layer}, we investigate the impact of integrating ingredient features at different layers of ResNet101 and InceptionV3. 
For ResNet101, integrating ingredient features after block2 achieves the best performance, while doing so after block4 significantly increases the average MAE. For InceptionV3, the best performance is observed when ingredient features are incorporated after the Mixed6e module and also fused in the auxiliary branch. The results show that injecting ingredient features in the intermediate layers achieves the best performance as intermediate layers retain semantic information while still preserving local details.

\begin{figure*}
  \centering
\includegraphics[width=0.95\textwidth]{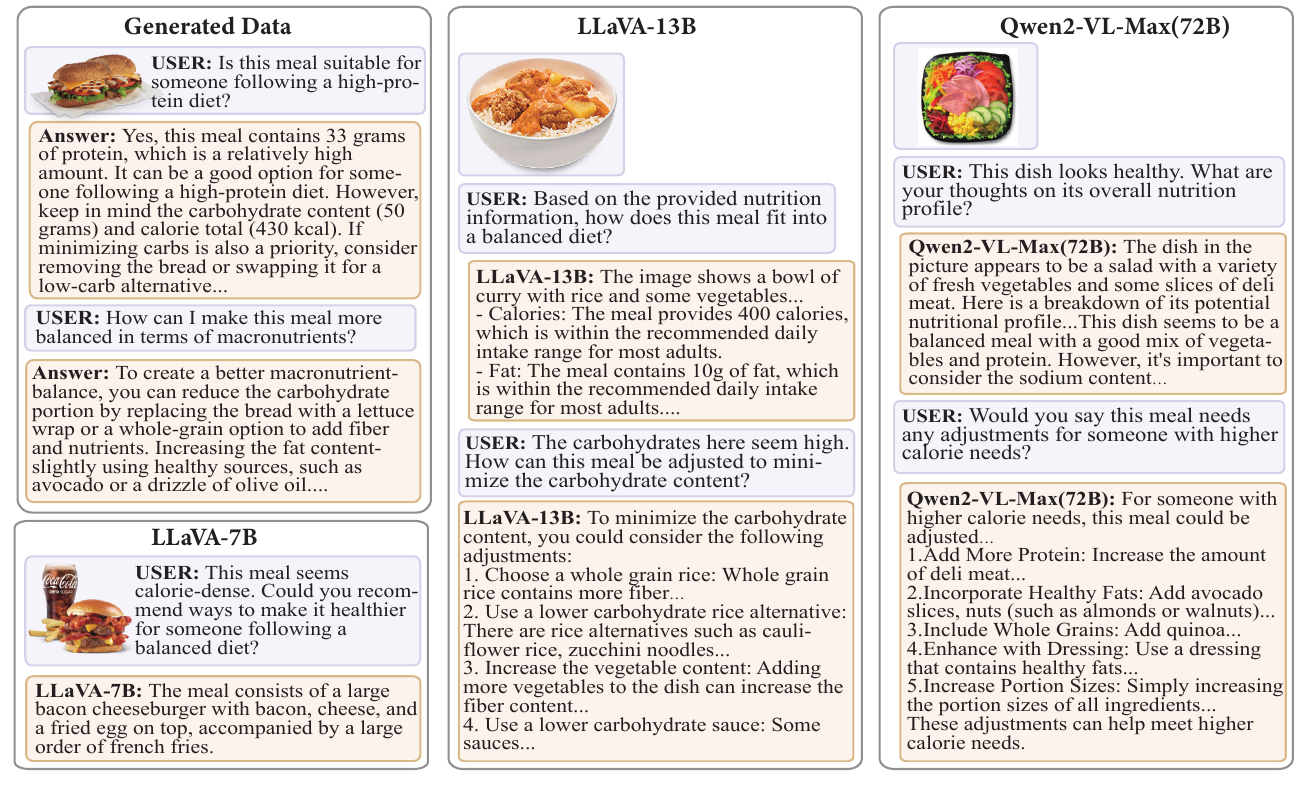} 
  \caption{Qualitative examples showcasing zero-shot diet analysis of multimodal large language models.}
  \label{fig:zero_shot}
\end{figure*}

\subsection{Zero-Shot Evaluation of Diet Analysis and Nutritional Reasoning}

\noindent \textbf{Threshold $\tau$ for augmented ingredient prediction.} 
Figure~\ref{fig:threshold} illustrates the effect of threshold for augmented ingredient prediction. 
As the threshold increases from 1 to 4, the average MAE decreases, reaching its lowest value of 61.26 at a threshold of 4. This indicates that an appropriate threshold effectively filters out irrelevant or noisy ingredients, enhancing nutrition prediction performance. However, when the threshold increases beyond 5, the average MAE begins to rise, suggesting that overly high thresholds may result in insufficient ingredient information, which compromises the model's generalization capability. 

\noindent \textbf{Large multimodal models.} To evaluate the impact of different large multimodal models (LMMs) on ingredient generation during inference, we utilize three LMMs during the testing phase with augmented ingredient prediction: LLaVA1.6 (7B)~\cite{liu2024improved}, Qwen2-VL (7B), and LLaVA-NeXT (Llama-3-8B)~\cite{zhang2024video}. Table~\ref{tab:ablation_LMM} shows that LLaVA1.6 achieves the best performance across all metrics. Qwen2-VL follows closely, showing slightly lower performance than LLaVA1.6 in Caloric and Fat but demonstrating improvements in Carb and Protein. In contrast, LLaVA-NeXT exhibited weaker overall performance, falling behind the other two models in all metrics.

Figure~\ref{fig:zero_shot} illustrates our generated data and evaluates the reasoning of diet-related questions answered by LLaVA(7B/13B) and Qwen2-VL-Max(72B) under zero-shot setting. In our generated data, the question-answer pairs are capable of providing accurate nutritional analysis and formulating reasonable diet suggestions based on the provided image. While clearly addressing users' needs, our QA pairs also consider the balance of other macronutrients (such as carbohydrates and fats) and offer specific suggestions for improving nutritional balance accordingly.

During testing, LLaVA-7B demonstrates a correct understanding of both the images and user queries, such as analyzing the high-calorie characteristics of the food. However, its responses lack specific support for nutrition content and do not adequate. For instance, when a query mentions that "high protein intake should be balanced with carbohydrates and fats," it fails to provide actionable suggestions. LLaVA-13B shows significant improvements over  LLaVA-7B in terms of the accuracy and interpretability of nutrition information. Its responses encompass a comprehensive analysis from total calorie content to macronutrient breakdown and offer specific recommendations to address user queries, reflecting greater practicality. Qwen2-VL-Max(72B) is capable of conducting a relatively comprehensive analysis too. Moreover, this model identifies potential issues in its responses, such as high sodium and calorie sources, and offers highly feasible and diverse solutions tailored to user requirements.

\section{Conclusion}
In this paper, we have presented a new dataset named FastFood 
with ingredient and nutrition annotations. 
Furthermore, we propose a model-agnostic method by integrating the ingredient and visual features for nutrition estimation. The results on both FastFood and Nutrition-5K show that our proposed method significantly outperforms the existing methods. In the future, we aim to explore the development of large multimodal models for advanced nutrition estimation and personalized diet suggestion generation.  

\section*{Acknowledgment}
This research/project is supported by the Ministry of Education, Singapore, under Academic Research Fund (AcRF) Tier 1 grant (No. MSS23C018) and Tier 2 (Proposal ID: T2EP20222-0046). Any opinions, findings and conclusions or recommendations expressed in this material are those of the authors and do not reflect the views of the Ministry of Education, Singapore.

\bibliographystyle{ACM-Reference-Format}
\bibliography{main}

\end{document}